\documentclass[%
 aip,
 amsmath,amssymb,
 reprint,%
]{revtex4-1}

\usepackage{graphicx}
\usepackage{dcolumn}
\usepackage{subfigure}
\usepackage{bm}

\usepackage[utf8]{inputenc}
\usepackage[T1]{fontenc}
\usepackage{mathptmx}
\usepackage{etoolbox}
\usepackage{float}
\usepackage{natbib}

\makeatletter
\def\@email#1#2{%
 \endgroup
 \patchcmd{\titleblock@produce}
  {\frontmatter@RRAPformat}
  {\frontmatter@RRAPformat{\produce@RRAP{*#1\href{mailto:#2}{#2}}}\frontmatter@RRAPformat}
  {}{}
}%
\makeatother
\begin{document}

\preprint{AIP/123-QED}

\title[PICL]{PICL: Physics Informed Contrastive Learning for Partial Differential Equations}

\author{Cooper Lorsung}
\affiliation{ 
Department of Mechanical Engineering, Carnegie Mellon University, 5000 Forbes Ave, Pittsburgh, PA 15213
}
 
\author{Amir Barati Farimani}
 \altaffiliation[Corresponding Author: ]{barati@cmu.edu}
\affiliation{ 
Department of Mechanical Engineering, Carnegie Mellon University, 5000 Forbes Ave, Pittsburgh, PA 15213
}
\affiliation{
Department of Biomedical Engineering, Carnegie Mellon University, 5000 Forbes Ave, Pittsburgh, PA 15213
}
\affiliation{
Machine Learning Department, Carnegie Mellon University, 5000 Forbes Ave, Pittsburgh, PA 15213
}

\date{\today}

\begin{abstract}
Neural operators have recently grown in popularity as Partial Differential Equation (PDE) surrogate models.
Learning solution functionals, rather than functions, has proven to be a powerful approach to calculate fast, accurate solutions to complex PDEs. 
While much work has been done evaluating neural operator performance on a wide variety of surrogate modeling tasks, these works normally evaluate performance on a single equation at a time.
In this work, we develop a novel contrastive pretraining framework utilizing Generalized Contrastive Loss that improves neural operator generalization across multiple governing equations simultaneously.
Governing equation coefficients are used to measure ground-truth similarity between systems.
A combination of physics-informed system evolution and latent-space model output are anchored to input data and used in our distance function.
We find that physics-informed contrastive pretraining improves accuracy for the Fourier Neural Operator in fixed-future and autoregressive rollout tasks for the 1D and 2D Heat, Burgers', and linear advection equations.
\end{abstract}

\maketitle

\section{Introduction}
\label{sec:sample1}
Contrastive learning frameworks have shown great promise in traditional machine learning tasks such as image classification\cite{pmlr-v119-chen20j, sohn_improved_2016}, with more recent works extending the applications to molecular property prediction\cite{Wang2022} and dynamical systems\cite{jiang2023training}.
While these works generally focus on classes that have distinct boundaries, weighted contrastive learning has been developed for cases where distinct samples are more similar to some samples than others.
This has been applied to visual place recognition\cite{GCL, Leyva-Vallina_2023_CVPR} as well as molecular property prediction\cite{imolclr}.
In the case of Partial Differential Equations (PDEs), weighted similarity can be useful in learning multiple operators simultaneously.
For example, a system governed by diffusion-dominated Burgers' equation behaves much more similarly to a system governed by the Heat equation than a system governed by advection-dominated Burgers' equation, and one model may be trained to learn both systems governed by the heat and Burgers' equations.

Many neural operators have been developed for various PDE surrogate modeling tasks \cite{li2023transformer, li2021fourier, Lu2021, patil2023hyena}.
Existing works aim to improve simulation speed through super-resolution \cite{shu_physics-informed_2023}, mesh optimization \cite{meshdqn}, and compressed representation \cite{hemmasian2023multiscale, romer, li2023latent, li2023scalable}.
Despite their promising results, few works have tested generalization across different operator coefficients for a single governing equation, or multiple governing equations.
Recently, Physics Informed Token Transformer\cite{lorsung2023physics} and PROSE\cite{liu2023prose} have shown promise in multi-system learning, and the CAPE module\cite{takamoto2023learning} has demonstrated the ability to incorporate different equation coefficients with good generalizability.
However, none of these works explicitly utilize the differences between systems, instead relying on a data driven approach to simultaneously learn the effect of equation coefficients and prediction.
A recent work, PIANO\cite{zhang2023deciphering}, uses operator coefficients, forcing terms, and boundary condition information for contrastive pretraining.
While promising, this work applies pretraining to varying systems with the same governing equation, rather than multiple governing equations.
Additionally, the PIANO framework does not take into account similarity between different, but similar, systems.
ConCerNet\cite{pmlr-v202-zhang23ao_concernet} also applies contrastive learning to the Heat equation and dynamical systems, but uses contrastive learning to minimize distance between embeddings from the same trajectory, rather than across different systems.

The aim of this work is to develop a contrastive framework that enables a single model to more effectively learn multiple operators by learning from the differences between systems explicitly.
This presents a number of challenges related to both the underlying mathematical theory, as well as practical implementation.
Namely, distance functions often utilize norms.
Euclidean distance, for example, utilizes the $L^2$ norm.
However, it is well known that differential operators are unbounded, and therefore have no norm.
Cosine similarity, another popular choice, is magnitude invariant, which can not capture different coefficient magnitudes.
Practically speaking, if we were to approximate our differential operators with finite difference matrices, we could use a matrix norm to easily define a distance function.
However, since we are using a single set of model weights to learn multiple operators, the matrix norm of model weights is the same for each operator, rendering the matrix norm useless for this case.
In this work, we develop a novel framework that overcomes these challenges.
Our contributions are as follows:
\begin{itemize}
    \item A novel similarity metric between PDE systems
    \item A novel neural operator based operator distance function
\end{itemize}
The similarity metric and distance function are combined using Generalized Contrastive Loss\cite{GCL, Leyva-Vallina_2023_CVPR} to form our framework: Physics Informed Contrastive Learning (PICL).
Our contrastive framework is benchmarked using the Fourier Neural Operator (FNO)\cite{li2021fourier} on popular 1D and 2D PDEs for fixed-future prediction and autoregressive rollout.
Further analysis also shows that models pretrained with PICL are able to clearly distinguish between different systems.
PICL shows significant improvement over standard training in 1D fixed-future and autoregressive experiments, with smaller improvement in 2D fixed-future and autoregressive experiments.
Latent space embeddings from PICL also show clustering of systems that behave similarly.

\section{Data Generation}
\subsection{1D Data}
In order to properly assess performance, multiple data sets that represent distinct physical processes are used.
In our case, we have the 1D Heat equation (eq. \ref{eq:Heat}), which is a linear parabolic equation, the linear advection equation (eq. \ref{eq:advection}) which is a linear hyperbolic equation, and Burgers' equation (eq. \ref{eq:Burgers}), and their 2D equivalents (eq. \ref{eq:2d_equations}).
All systems have periodic boundary conditions.
Adapting the setup from \cite{brandstetter2023message}, we generate the 1D data for the homogeneous Heat and Burgers' equations, with linear advection data being generated analytically.
\begin{equation}
    \partial_t u - \beta\partial_{xx}u = 0
\label{eq:Heat}
\end{equation}
\begin{equation}
    \partial_t u + \gamma\partial_{x}u = 0
\label{eq:advection}
\end{equation}
\begin{equation}
    \partial_t u + \alpha\partial_{x}u^2 - \beta\partial_{xx}u = 0
\label{eq:Burgers}
\end{equation}
In this case, a large number of sampled coefficients allow us to generate data for many different systems.
We use multiple parameters for each equation to generate our diverse data set, given by: $\alpha \in \left\{0.5, 1.0, 2.0, 5.0\right\}$, $\beta \in \left\{0.2, 0.5, 1.0, 2.0, 5.0\right\}$, and $\gamma \in \left\{0.5, 1.0, 2.0, 5.0\right\}$.

The initial condition is given by:
\begin{equation}
    u\left(x\right) = \sum_{j=1}^JA_j \sin\left(\frac{2\pi l_j x}{L} + \phi_j\right) \\
    \label{eq:1d_ic}
\end{equation}
The parameters $A$, $l$, and $\phi$ can be sampled to give us many initial conditions for each set of system coefficients.
In our case, $J=5$ and our system size is $L=128$.
The parameters are sampled as follows: $A_j \sim \mathcal{U}(-0.5, 0.5)$, $l_j \sim \{1, 2, 3\}$, $\phi_j \sim \mathcal{U}(0, 2\pi)$.
We generated different data sets for each of pretraining, fine-tuning/standard training, validation, and evalutation.
Each dataset has different initial conditions sampled from the same distributions.
Our system was spatially discretized with 200 evenly spaced points and temporally discretized with 200 evenly spaced samples in time.
During training both our spatial and temporal discretizations were evenly downsampled to 50 points.

\subsection{2D Data}
Our 2D data is generated from the 2D homogeneous Heat, Burgers, and Advection equations over a the domain $[x,y] = [-1,1]^2$ for 32 timesteps with $\Delta t = 0.02$ on 64x64 grid that is evenly downsampled to 32x32 for pretraining, fine-tuning, and evaluation.
We use finite-differences for 2D data generation adapted from \cite{Barba2019}.
\begin{equation}
    \begin{split}
    \partial_tu - \beta\nabla^2u &= 0 \\
    \partial_tu + \mathbf{c}\cdot\nabla u &= 0 \\
    \quad \partial_tu + u\left(\mathbf{c}\cdot\nabla u\right) &- \beta\nabla^2u = 0
    \end{split}
    \label{eq:2d_equations}
\end{equation}
We use existing 2D data sets from Zhou et. al.\cite{zhou2024strategies_pretrainingpdes}, where operator coefficients are sampled uniformly from $\beta \in \left[0.02, 0.03\right]$ for the Heat equation, $\mathbf{c} = c_{x,y}\in\left[2.5,3.0\right]^2$ for the advection equation, and $\beta \in \left[0.005, 0.0075\right]$ and $\mathbf{c} = c_{x,y} \in \left[1.0, 1.25\right]^2$ for Burgers equation.
Operator coefficient distributions were chosen so that system evolution was of approximately the same magnitude over the temporal window.
The initial condition is sampled from equation \ref{eq:2d_ic}.
For each initial condition, the coefficients are sampled from: $A_j\in\left[-0.5, 0.5\right]$ , $\omega_j\in\left[-0.4,0.4\right]$, $l_{xj}l_{yj} \in \{1,2,3\}$, and $\phi_j \in \left[0, 2\pi\right)$.
We use fixed values $J=5$ and $L=2$.
\begin{equation}
    u\left(x\right) = \sum_{j=1}^JA_j \sin\left(\frac{2\pi l_xj x}{L} + \frac{2\pi l_{jy}}{L}\phi_j\right) \\
    \label{eq:2d_ic}
\end{equation}
Our 2D pretraining, fine-tuning/standard training, and evaluation sets have different initial conditions and coefficients sampled from the same distributions.

\section{Method}

The goal of neural operator learning is to learn the mapping $G_{\theta}: \mathcal{A} \to \mathcal{S}$, parameterized by $\theta$, from input function space $\mathcal{A}$ to solution function space $\mathcal{S}$\cite{JMLR:v24:21-1524}.
When looking at specific functions, we can view our operator as acting on a specific input function, $a$, and mapping it to a specific output function $s$, as: $\mathcal{G}_{\theta}(a) \to s$.
Specifically, we are learning various operators $\mathcal{G}_\theta$, and pretraining in the neural operator latent embeddings, represented by $\mathcal{G}'_{\theta}$.
However, as previously mentioned, our single set of model weights $\theta$ acts as different operators depending on the input data.
While we cannot practically or mathematically utilize the individual operators themselves for our similarity metric or distance function, we can utilize the \textit{effect} the operators have on our system.

\subsection{Generalized Contrastive Loss}
In this work, we use the Generalized Constrastive Loss (GCL)\cite{GCL, Leyva-Vallina_2023_CVPR}, given below in equation \ref{eq:GCL}.
\begin{equation}
    \begin{split}
    \mathcal{L}_{GCL}(z_i, z_j) =& \psi_{i,j} \frac{d(z_i, z_j)^2}{2} \\
    & + \left(1 - \psi_{i,j}\right)\frac{\text{max}(\tau - d(z_i, z_j),0)^2}{2}
    \end{split}
\label{eq:GCL}
\end{equation}
The first term aims to minimize distance between similar samples.
In the second term, $\tau$ acts as a margin, above which samples are considered to be from different systems. 
The second term therefore maximizes distance between unlike samples that are below the margin threshold.
The key components of this loss function are the distance function between samples, $d(z_i, z_j)$, and the similarity metric, $\psi_{i,j}$.
While the distance is calculated model with output, known properties from our system are used in the similarity metric.
In this case, we use operator coefficients to measure similarity.

\subsection{Similarity Metric}
We generate multiple trajectories for each combination of equation parameters: $\alpha$, $\beta$, and $\gamma$, which can be stored in a vector as $\theta = \left[\alpha, \beta, \gamma\right]$ for 1D equations, and $\theta = \left[\left\lVert \left[a_x, a_y\right] \right\rVert_2 \nu, \left\lVert \left[c_x, c_y\right] \right\rVert_2\right]$, for Burgers advection coefficients $a_{x,y}$, and linear advection coefficients $c_{x,y}$.
These parameters select which governing equation is being used as well as the governing equation properties.
Once we have constructed the weight vector for each system, the novel magnitude-aware cosine similarity is used to calculate similarity between our weight vectors.
\begin{equation}
    \psi(\theta_i, \theta_j) = \frac{\sqrt{\left|\theta_i \cdot \theta_j\right|}}{\text{max}\left( \left\lVert \theta_i \right\rVert_2, \left\lVert \theta_j \right\rVert_2 \right)}
\end{equation}
While similar to cosine similarity, taking the maximum of both input vectors normalizes the output to 1 if the magnitude of the dot product is equal to the magnitude of the larger vector, i.e. the inputs are identical.
Magnitude-awareness is critical for PDEs, because, for example, a highly diffusive Heat system behaves differently form a weakly diffusive Heat system. 

\subsection{Physics Informed Distance Metric}
Measuring distance in latent space plays a vital role in contrastive learning.
In many cases, Euclidean distance or cosine similarity are used.
However, in the case of operator learning for PDEs, it is well known that differential operators are unbounded, and therefore a metric cannot be defined.
With this in mind, a measure of distance must utilize additional information outside of the analytical governing equations.
When we have different initial conditions for a given system, i.e. different initial sine waves evolving according to the heat equation with diffusion coefficient of 1, we expect that the system evolution will look similar between these different initial conditions.
That is, the difference between the first and second frames of these two systems should show a more similar evolution than if one of the systems evolved according to the Advection equation.
This distance, which we call the system distance, is given in equation \ref{eq:dsys}.
\begin{equation}
    d_{system}(u_i, u_j) = u_i^{t+1} - u_j^{t} 
 \label{eq:dsys}
\end{equation}

\begin{figure}[h]
    \centering
    \includegraphics[width=\linewidth]{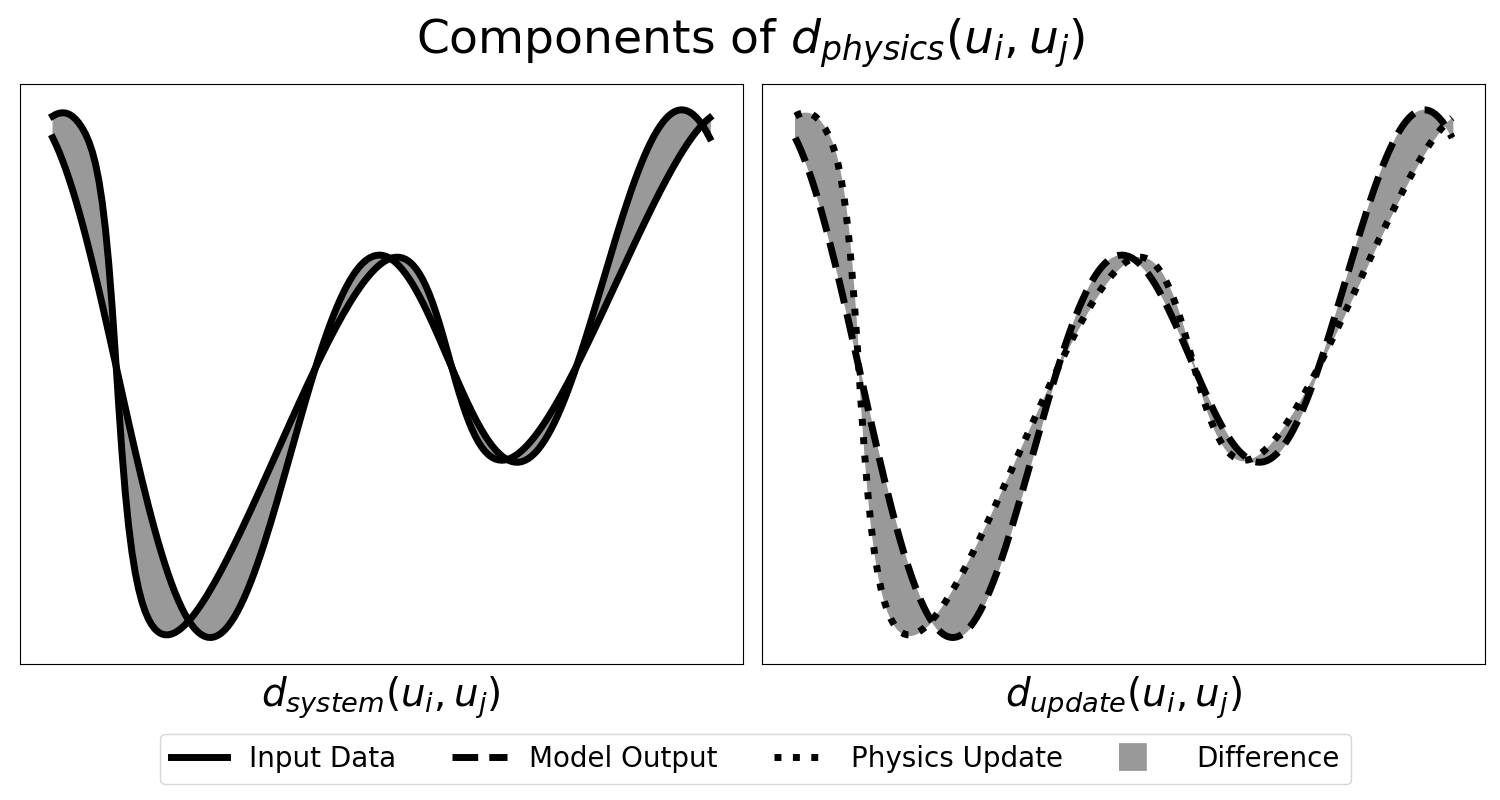}
    \caption{Our pretraining distance function measures distance between the difference of successive frames of input data, and the difference of model output and model output with a physics-informed update.}
    \label{fig:d_physics}
\end{figure}

Since our models are learning the operators themselves, we must also utilize model output so that errors can be backpropagated.
Similar to system difference, we can calculate distances between our predicted states, with additional physics information to calculate the next step, given in equation \ref{eq:dupdate_1}.
\begin{equation}
    d_{update}(u_i, u_j) = F(G_{\theta}(u_i^t)) - G_{\theta}(u_j^t)
    \label{eq:dupdate_1}
\end{equation}
where $G_{\theta}(u_i)$ is our parameterized model and $F(\cdot)$ is our numerical update operator.
In our case, at timestep $t$, 
\begin{equation}
    F(z^t) = z^t + 2\alpha_z z\partial_xz^t - \beta_z \partial_{xx} z^t + \gamma_z \partial_{x} z^t = z^{t+1}
\end{equation}
for $z = G_{\theta}(u)$.
Each differential operator is calculated with a finite difference approximation given in appendix \ref{sec:physics_informed_updates}.
This distance should be similar to $d_{system}$, since they are both after a single timestep, and so we anchor $d_{update}$ to $d_{system}$, inspired by triplet loss\cite{Schroff_2015_CVPR}, given in equation \ref{eq:picl_distance}.
Anchoring is done so that our system update distances are not minimized to 0, which does not accurately reflect the effect of the operators.
While this formulation of $d_{system}$ relies on multiple snapshots as input data, we can extend the functionality to only using initial conditions as input by replacing $u_{i}^{t+1}$ with $F(u_{u}^{t})$.
Incorporating physics information here serves two purposes.
First, it helps smooth our model output. A very jagged model output would result in large derivatives that would be very different from our input system evolution.
Second, it enforces that our predicted states have a similar numerical update to our input data, which helps ensure the representation is similar to our input data.
The components of $d_{physics}$ are given in figure \ref{fig:d_physics} for the distance between a sample and itself.

\begin{equation}
    \begin{split}
        d_{physics}(u_i, u_j, z_i, z_j) =& \left\lVert d_{system}(u_i, u_j) - d_{update}(z_i, z_j)\right\rVert^2 \\
    \end{split}
    \label{eq:picl_distance}
\end{equation}

\subsection{Training Procedure}
\begin{figure}[h]
    \centering
    a) Pretraining Procedure\\
    \begin{subfigure}
        \centering
        \includegraphics[width=\linewidth]{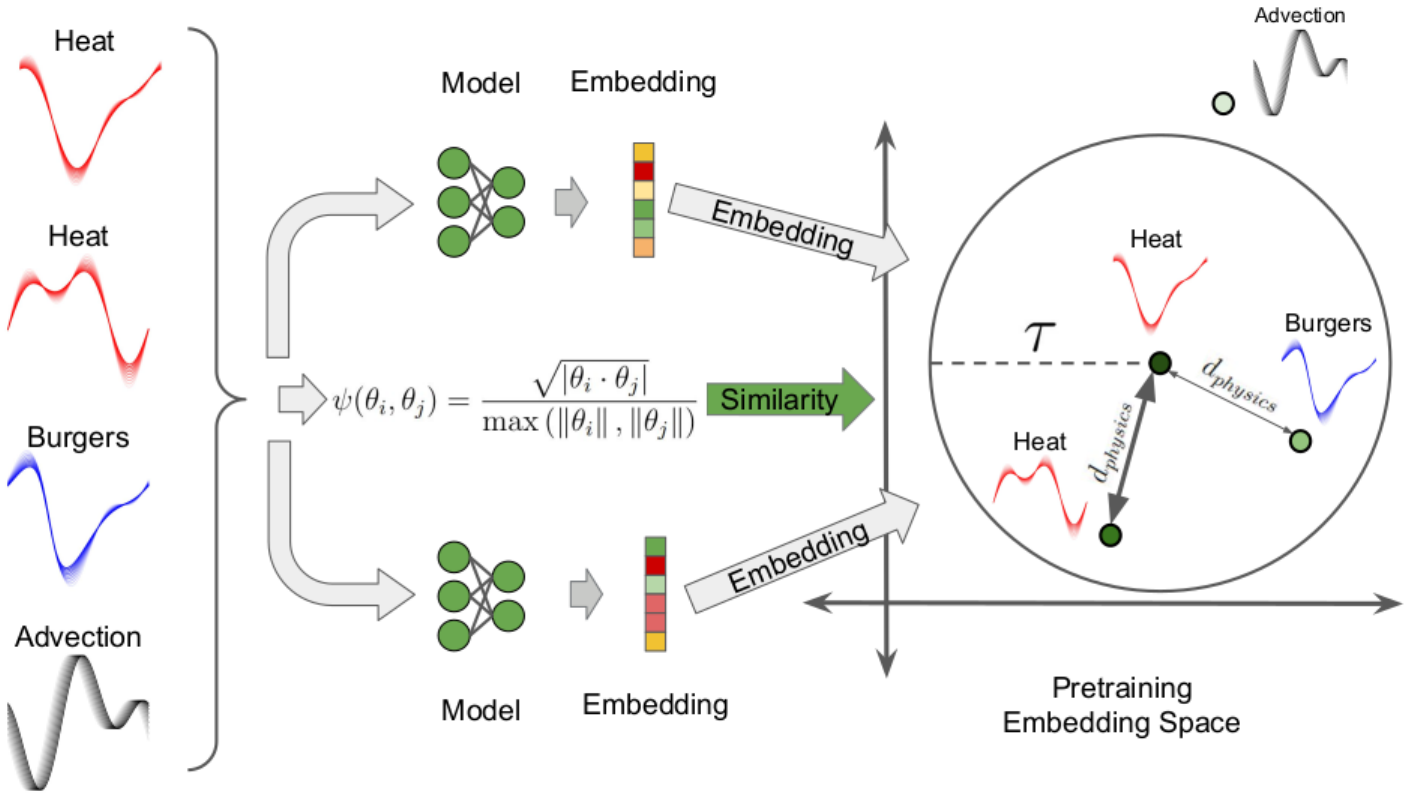}
    \end{subfigure}
    b) Fine-tuning Procedure\\
    \begin{subfigure}
        \centering
        \includegraphics[width=0.6\linewidth]{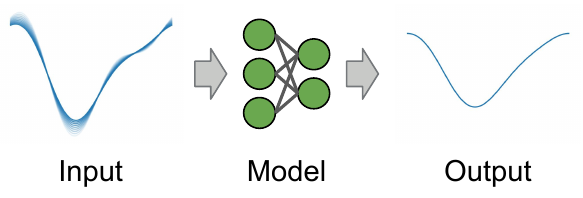}
    \end{subfigure}
    \caption{Our two-step training procedure first pretrains on all three equations simultaneously (a), then fine-tunes on each equation individually (b). Darker green points represent higher similarity to our Heat sample. Bolder arrows represent stronger attraction between samples in embedding space.}
\label{fig:training_procedure}
\end{figure}

We employ a two step training procedure seen in figure \ref{fig:training_procedure}, where we first use contrastive pretraining and then standard training.
In both stages, we train our model end-to-end.
During pretraining, we use our PICL contrastive loss function.
After pretraining, during fine-tuning, we use a standard training procedure.
We do not employ weight-freezing after pretraining because we have empirically found this to have lower predictive accuracy.
The training procedure is given in figure \ref{fig:training_procedure}.

%

\section{Results}
PICL is now benchmarked against standard training methods on our 1D and 2D data sets.
Data is split such that trajectories are not used for both training and evaluation.
In each experiment, five random seeds and model initializations were used.
Reported values are the mean and standard deviation relative $L^2$ norm\cite{li2021fourier} of these five runs.
For all experiments, we compare PICL-pretrained FNO against standard FNO for each of the Heat, Burgers', and Advection equations individually.
In the 1D case, we compare against all equations in a combined data set as well.
For each experiment, we pretrain using the combined data set, and use the learned model weights after pretraining for fine-tuning.
In 1D and 2D, we use 5000 and 3072 samples from each equation in pretraining, respectively.
Hyperparameters for each experiment are given in Appendix \ref{sec:hyperparams}.
We have found that using a one cycle learning rate scheduler leads to improved performance over the standard step learning rate scheduler.
Further benchmarking of PICL was done in in Zhou et al.\cite{zhou2024strategies_pretrainingpdes}, where UNet\cite{ronneberger2015unetconvolutionalnetworksbiomedical}, DeepONet\cite{Lu_2021_deeponet}, OFormer\cite{li2023transformer} where used for various pretraining strategies for 2D in-distribution and out-of-distribution experiments, including Navier-Stokes data.
Data augmentation was also used to more fully explore existing methods in the PDE surrogate modeling space.
PICL tends to improve performance on that set of experiments, and transfer learning tends to improve performance a bit further.

\subsection{1D Results}
To construct our train set, we sample the specified number of samples per coefficient combination.
That is, for two samples per coefficient combination of Burgers' data, we have two samples with $\alpha=0.2$, $\beta=0.2$, two samples with $\alpha=0.2$, $\beta=0.5$, etc.

\subsubsection{Fixed Future}
\label{sec:1d_ff}
\begin{figure}
    \centering
    \includegraphics[width=\linewidth]{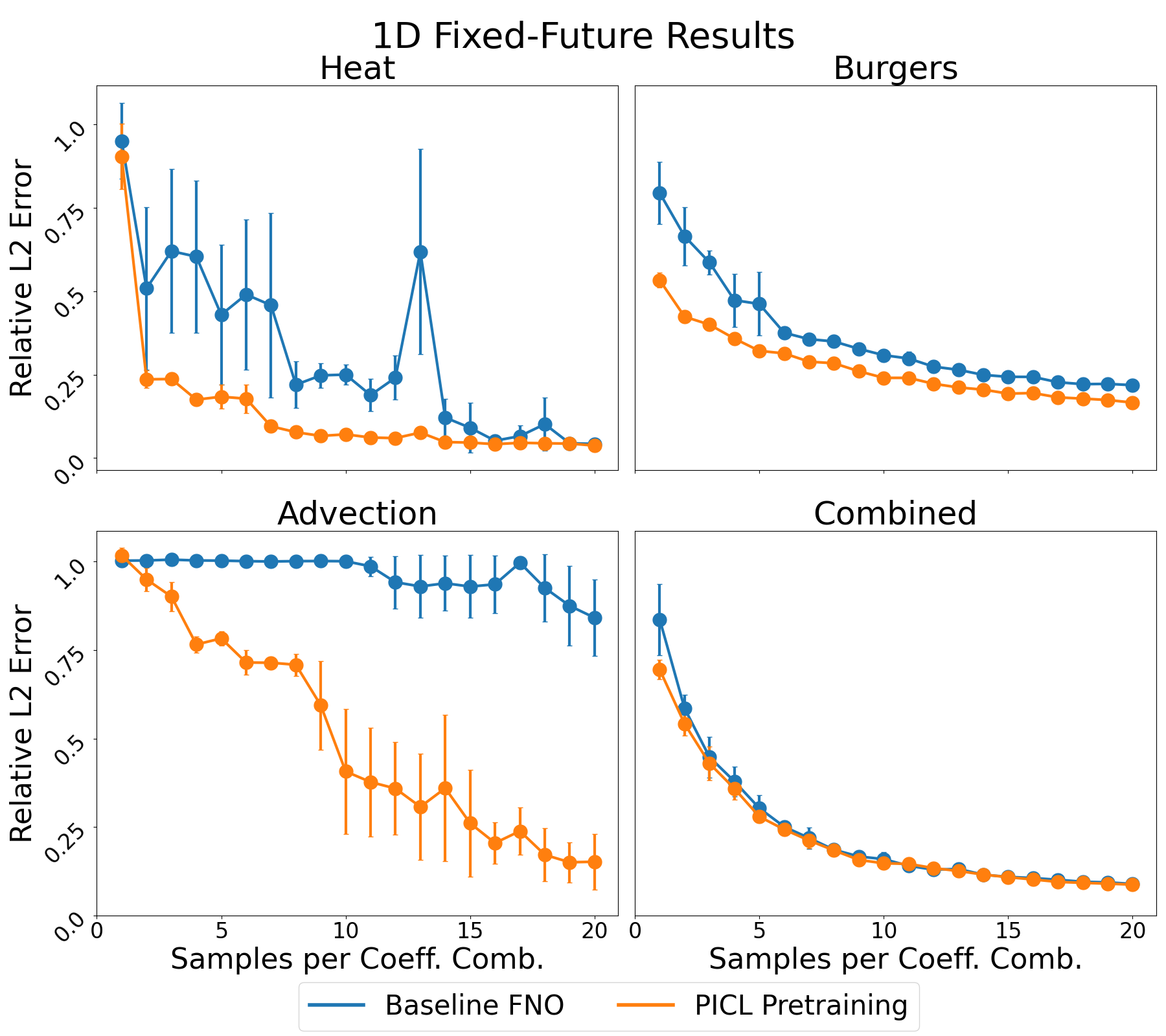}
    \caption{1D comparison of fixed-future performance between FNO and FNO pretrained using PICL.}
    \label{fig:1d_ff_comp}
\end{figure}

In this experiment we aim to learn a mapping from the initial condition given equation coefficients and target time, to a fixed time in the future the operator $\mathcal{G}_{\theta}: a(\cdot, t_i)|_{i=0} \to u(\cdot, t_i)|_{i=49}$ for $\Delta t = 0.0603$.
Results are given in figure \ref{fig:1d_ff_comp}.
In this case, early stopping was used where the model weights with best performance on the validation set were used for evaluation on the test set.
We see that PICL shows significant improvement over standard training.
Further comparison against a pretraining approach the does not use physics information is given in Appendix \ref{app:passthrough_tsne}.

\subsubsection{Autoregressive Rollout}
\begin{figure}[h]
    \centering
    \includegraphics[width=\linewidth]{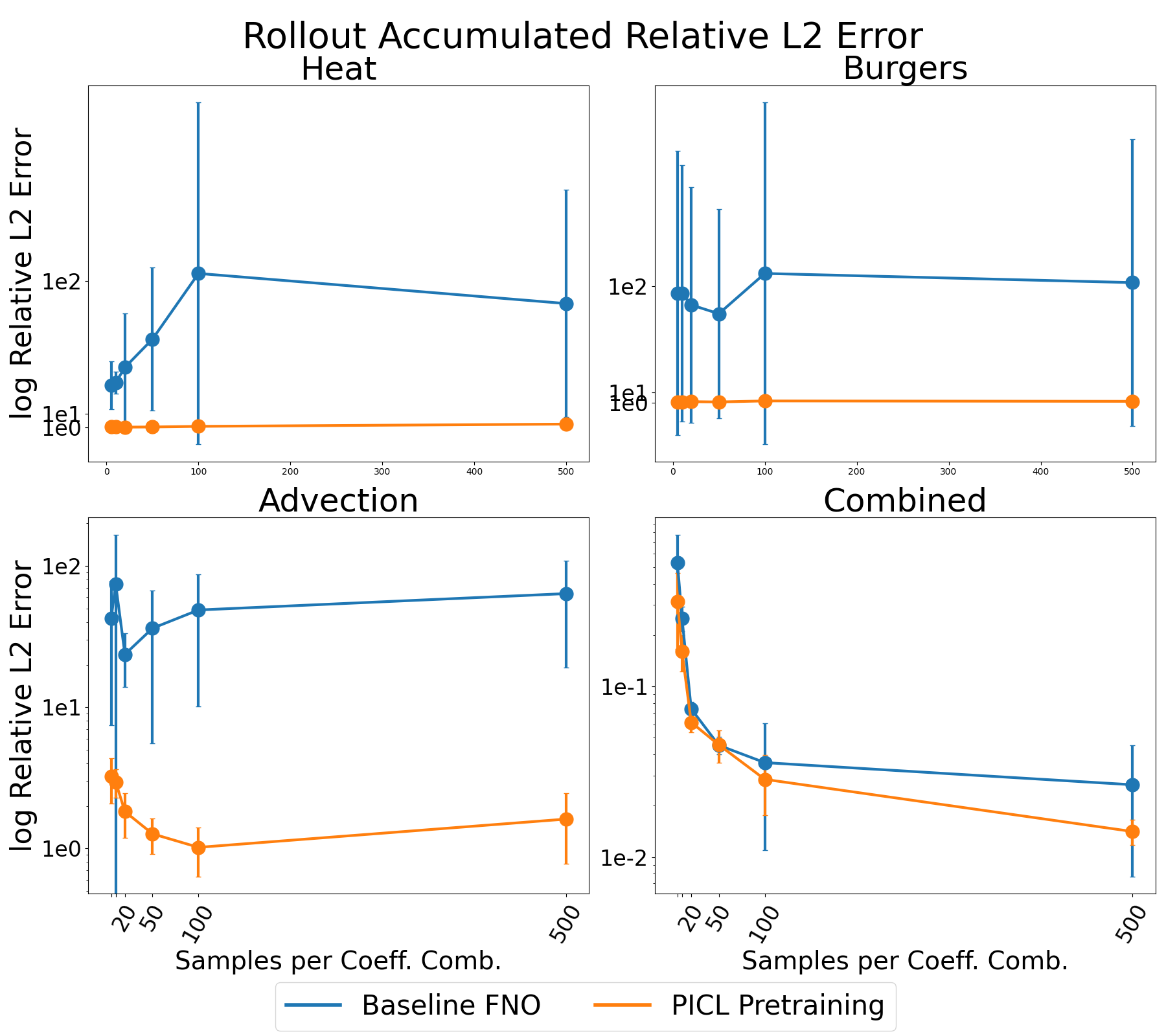}
    \caption{Comparison of autoregressive rollout performance between FNO and FNO pretrained using PICL.}
    \label{fig:acc_rollout}
\end{figure}

To test autoregressive rollout, we train each model given a frame of data, the equation coefficients, and the target time to predict the next step.
That is, we are learning the operator $\mathcal{G}_{\theta}: a(\cdot, t_i)|_{i=n} \to u(\cdot, t_i)|_{i=n+1}$, again with $\Delta t = 0.0603$.
After training, we test rollout by using the initial condition as input, predicting the next step, then using the predicted frame $\tilde{u}(\cdot, t_1)$ to predict frame $2$, etcetera, until we reach the full trajectory, as in Brandstetter et al.\cite{brandstetter2023message}.
Total accumulated error is given in figure \ref{fig:acc_rollout}, where we see 2 order of magnitude improvement over baseline when using PICL for individual data sets that is maintained even with more fine-tuning data, and improvement that is growing with number of samples in our combined data set.
Next-step predictive performance and autoregressive rollout plots are given in plots \ref{fig:1d_next_step_pred} and \ref{fig:rollout}, respectively.
Comparison of next-step predictive performance, and plots of autoregressive error are given in Appendix \ref{app:ns_ar}.

\subsection{2D Results}
We use random sampling from our fine-tune and test sets, and use early stopping by reporting best performance on our test set during training.

\subsubsection{Fixed-Future}
For fixed-future training, we use the initial condition to predict the final state.
We see improvement over baseline in table \ref{tab:2d_ff} for Heat and Burgers equations when using PICL pretraining.
\begin{table}[h]
    \centering
    \begin{tabular}{c|cccc}
         Model & Heat & Burgers' & Advection \\
         \hline
         FNO & 2.21 $\pm$ 0.06 & 4.33 $\pm$ 0.24 & 72.32 $\pm$ 0.30 \\
         FNO Pretrained & \textbf{1.85 $\pm$ 0.18} & \textbf{4.13 $\pm$ 0.09} & 72.38 $\pm$ 0.31 \\
    \end{tabular}
    \caption{Fixed-Future 500 samples/equation Relative $L^2$ Norm ($\times 10^{-2}$)}
    \label{tab:2d_ff}
\end{table}

\subsubsection{Autoregressive Rollout}
In autoregressive rollout we use four frames of data as input, four frames as the temporally bundled output, and one pushforward step\cite{brandstetter2023message}.
Reported values are total accumulated error over autoregressive rollout for the entire trajectory, averaged over our five random seeds.
We see that we get improvement over baseline with PICL pretraining in table \ref{tab:2d_ar} across all of our data sets.
\begin{table}[h]
    \centering
    \begin{tabular}{c|cccc}
         Model & Heat & Burgers' & Advection \\
         \hline
         FNO & 0.388 $\pm$ 0.029 & 1.078 $\pm$ 0.066 & 6.736 $\pm$ 0.054 \\
         FNO Pretrained & \textbf{0.378 $\pm$ 0.01} & \textbf{1.032 $\pm$ 0.020} & \textbf{6.688 $\pm$ 0.132}\\
    \end{tabular}
    \caption{Autoregressive Rollout Error Accumulation 500 samples/equation Relative $L^2$ Norm}
    \label{tab:2d_ar}
\end{table}

\section{Discussion}
PICL shows improvement in FNO generalization.
In our 1D experiments, for both Heat and Burgers', we use the 5th-order accurate WENO5 method for data generation.
For advection, we use the analytical solution.
In all cases, we use a numerical update scheme that is of lower order accuracy.
For Heat, we use standard the 2nd-order centered difference scheme in space, and first order backward difference in time.
For Burgers', we use the same 2nd-order centered difference scheme in space for the diffusion term, and the second-order upwind scheme in space for the nonlinear advection term, coupled with a first-order backward difference scheme in time.
Lastly, for the linear advection, we use the same second-order upwind in space and first-order backward difference in time.
Despite this, PICL significantly improves results over standard training for fixed-future experiments and autoregressive rollout experiments in 1D.
Additionally, PICL is robust to stability constraints. In our 1D case, we use a timestep of $\Delta t = 0.6030$, with a spatial discretization of $\Delta x = 2.5729$ after downsampling.
For our upwind scheme, this gives us a CFL number of $\frac{5\Delta t}{\Delta x} = 1.17$, larger than the stability constraint of 1.
For our diffusion scheme, we obey the stability constraint of $\frac{5\Delta t}{\Delta x^2} = 0.455 < 0.5$.
Finally, in our 2D case we see improvement over standard training for both fixed-future prediction and autoregressive rollout despite continuous and significantly smaller coefficient distributions, making each system very similar.
Similar systems are more challenging to distinguish between, making the pretraining task more difficult.
Despite this, PICL is able to improve performance up to $16\%$ in our fixed-future experiment, and $4\%$ in our autoregressive experiment.

\begin{figure}[h!]
    \centering
    \includegraphics[width=\linewidth]{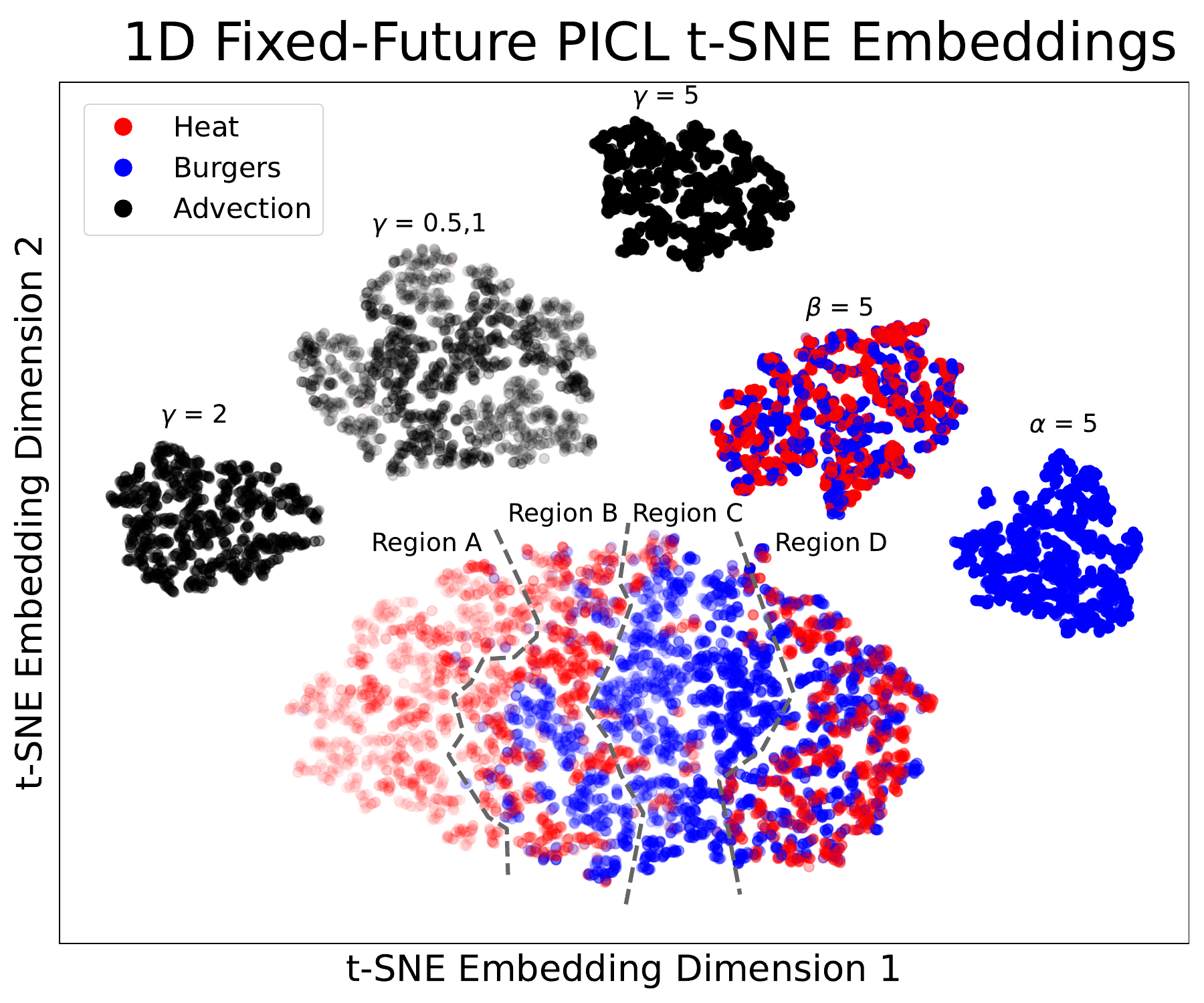} \\
    \caption{t-SNE of latent embeddings after PICL pretraining. We see clear clustering of similar systems, denoted by color and transparency. Advection systems are clustered separately from Heat and Burgers systems, strongly diffusive systems are clustered, strongly advective Burgers systems are clustered, and weakly to moderately diffusive and advective systems are clustered.}
    \label{fig:tsne}
\end{figure}

\begin{figure}[h!]
    \centering
    \includegraphics[width=\linewidth]{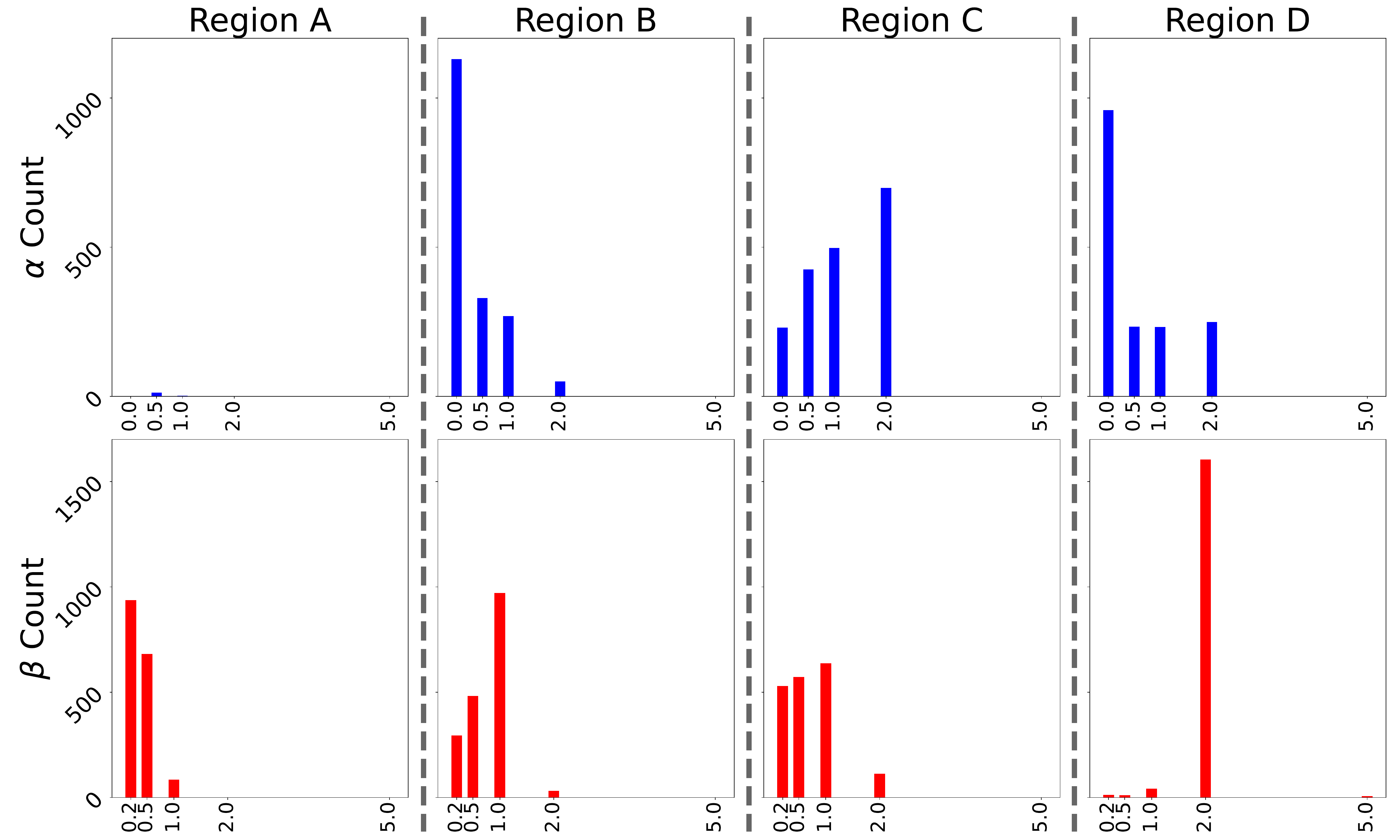}
    \caption{Distribution of coefficients in highlighted regions. We see Region A contains weakly diffusive systems, Region B contains weakly diffusive Heat and Burgers systems, Region C contains moderately advective Burgers systems, and Region D contains moderately advective Heat and Burgers systems.}
    \label{fig:parameter_distribution}
\end{figure}

We also check how well PICL allows our model to learn difference between systems by analyzing t-SNE embeddings of our latent representations in figure \ref{fig:tsne}.
Heat systems are given by red points, Advection systems are given by black points, and Burgers systems are given by blue points.
Parameter distributions for regions A through D are given in figure \ref{fig:parameter_distribution}.
We have weakly diffusive heat emeddings in region A, systems emeddings that are weakly diffusive and weakly advective in region B, moderately advective embeddings in region C, and moderately diffusive systems in region D, broadly going from smaller to larger coefficients as we go left to right.
In the advection clusters, we have 100\% of $\gamma = 5$ embeddings in the corresponding cluster, 99.5\% of $\gamma = 2$ embeddings in the corresponding cluster, and 100\% of the $\gamma = 0.5$ and 99.5\% of the $\gamma = 1$ embeddings in the corrsponding cluster.
In the $\beta = 5$ cluster, we have 87.1\% of embeddings with that $\beta$ values, for both Heat and Burgers' systems.
Lastly, we have 99.3\% of $\alpha = 5$ embeddings in the corrsponding cluster.
The t-SNE plot matches our broad intuition that diffusion dominated Burgers systems behave similarly to strongly diffusive Heat systems, advection dominated Burgers systems behave similarly to each other, and advection systems behave differently from Heat and Burgers systems.
More subtley, we have weakly diffusive Heat embeddings clustered closely, but separated from weakly diffusive Burgers embeddings, moderately advective Burgers embeddings clustered closely, but separated from weakly advective Burgers embeddings.
Overall, our embeddings match both broad and subtle intuition excellently.
The t-SNE plot from passthrough pretraining is given in Appendix \ref{app:passthrough_tsne}, where we see distinct clusters for each combination of operator coefficients.
While this does learn the underlying structure of the data based solely on operator coefficients, it does not match our understanding of the physics, like with PICL pretraining.




\section{Conclusion}
PICL offers a novel physics informed contrastive framework that improves FNO downstream performance on 1D and 2D homogenous PDE systems.
PICL leverages physics informed updates by anchoring predicted state updates to input data updates.
Our framework utilizes magnitude-aware cosine similarity to measure similarity between physical systems, which addresses mathematical limitations in operator theory.
Additionally, our distance function measures the distance between model outputs, addressing the challenge of measuring distance between different systems with a single set of model weights.
Combining our distance function with physics-informed updates enforces that our model output evolves similarly over time for similar systems, and that evolution behaves according to our known governing equations.
The drawbacks of PICL are that additional compute is required for pretraining, and governing equations need to be known.
PICL is currently only applicable when governing equation information is exactly known.
Future works include developing strategies to incorporate both static and time-dependent forcing terms to our distance function.
Applying PICL to more complex 2D and 3D systems, and a broader array of equations are also areas of interest.
Higher-dimensional systems offer an additional challenge that the magnitude of our distance function can varies significantly more than in our current experiments.
This, in turn, makes learning the differences between the effects of various operators more challenging.

\section{Acknowledgements}
CL would like to thank Anthony Zhou for providing the 2D data.
This material is based upon work supported by the National Science Foundation under Grant No. 1953222.

\section{Author Declarations}
The authors have no conflicts of interest to discose.

\section{Author Contributions}
\textbf{Cooper Lorsung}: Conceptualization (equal), Data Curation (lead), Formal Analysis (lead), Investigation (lead), Methodology (lead), Software (lead), Validation (lead), Visualization (lead), Writing/Original Draft Preparation (lead), Writing/Review \& Editing (equal).
\textbf{Amir Barati Farimani}: Conceptualization (equal), Funding Acquisition (lead), Methodology (supporting), Project Administration (lead), Supervision (lead), Validation (supporting), Writing/Review \& Editing (equal)

\section{Data Availability}
All code and data will be available at \href{https://github.com/CoopLo/PICL}{https://github.com/CoopLo/PICL}.

\newpage

\newpage
\clearpage
 \bibliographystyle{elsarticle-num} 
 \bibliography{the}

\appendix

\section{Physics-Informed Updates}
\label{sec:physics_informed_updates}
For each operator we use a low-order finite-difference scheme to calculate our updates.
In each equation $i$ represents the spatial coordinate.
For both our linear and nonlinear advection operators we use the upwind scheme since our advection velocity is always positive.
Our 1D finite-difference schemes are given below:
\begin{equation}
    \partial_{x}u = \frac{\Delta t}{\Delta x}\left(u_{i} - u_{i-1}\right)
\end{equation}
\begin{equation}
    \partial_{xx}u = \frac{\Delta t}{\Delta x^2}\left(u_{i+1} - 2u_{i} + u_{i-1}\right)
\end{equation}
\begin{equation}
    u\partial_{x}u = \frac{\Delta t}{\Delta x}u_{i}\left(u_{i} - u_{i-1}\right)
\end{equation}
Our 2D schemes extend the 1D case, following \cite{Barba2019}, and are given below.
For our state $u_{i,j}$, $i$ represents the x-coordinate and $j$ represents the y-coordinate.
\begin{equation}
    \nabla u = \Delta t \left[\frac{u_{i,j} - u_{i-1,j}}{\Delta x} + \frac{u_{i,j} - u_{i,j-1}}{\Delta y}\right]
\end{equation}
\begin{equation}
    \begin{split}
    \nabla^2u =& \Delta t\left[\frac{u_{i+1,j} - 2u_{i,j} + u_{i-1,j}}{\Delta x^2}\right. \\
    & \left.+ \frac{u_{i,j+1} - 2u_{i,j} + u_{i,j-1}}{\Delta y^2}\right]
    \end{split}
\end{equation}
\begin{equation}
    u\nabla u = \Delta t u_{i,j}\left[\frac{u_{i,j} - u_{i-1,j}}{\Delta x} + \frac{u_{i,j} - u_{i,j-1}}{\Delta y}\right]
\end{equation}
\section{Experiment Hyperparameters}
\label{sec:hyperparams}
The OneCycle learning rate scheduler was used for all experiments.
Model architecture and training hyperparameters were hand-tuned, with an emphasis on keeping hyperparameters between baseline and PICL pretraining as similar as possible.
Choosing $\tau$ to be the same order of magnitude as $d_{physics}$ from an untrained model is a good starting point for further tuning, where $\tau = $ mean when a good numerical value of $\tau$ could not be found.

\subsection{1D Hyperparameters}
For all 1D experiments, we used a 1D FNO with hidden width of 32 and 8 modes and 3 layers.
We trained for 500 epochs in the fixed-future experiment and 100 epochs in the autoregressive experiment.
Pretraining was done for 20 epochs in the fixed-future experiment and 5 for the autoregressive rollout experiment, with 3072 samples from each equation.
\subsubsection{Fixed-Future Hyperparameters}
\label{sec:ff_hyperparams}
Hyperparameters for finetuning and baseline training are given in table \ref{tab:1d_ff_train}, and for pretraining in table \ref{tab:1d_ff_pre}.
\begin{table}[H]
    \caption{Fine-tuning and Baseline Hyperparameters}
    \centering
    \begin{tabular}{c|cccccc}
         Model & Batch Size & Learning Rate & Weight Decay & Dropout & \\
         \hline
         FNO & 32 & 1E-3 & 1E-4 & 0.0 \\
         PICL FNO & 32 & 1E-3 & 1E-4 & 0.0 \\
    \end{tabular}
    \label{tab:1d_ff_train}
\end{table}
\begin{table}[H]
    \caption{Pretraining Hyperparameters}
    \centering
    \resizebox{\linewidth}{!}{
    \begin{tabular}{c|ccccccc}
         Model & Batch Size & Learning Rate & Weight Decay & Dropout & $\tau$ \\
         \hline
         Pretrain FNO & 512 & 1E-2 & 1E-8 & 0.00 & 5 \\
    \end{tabular}}
    \label{tab:1d_ff_pre}
\end{table}

\subsubsection{Autoregressive Rollout Hyperparameters}
Hyperparameters for finetuning and baseline training are given in table \ref{tab:1d_ar_train}, and for pretraining in table \ref{tab:1d_ar_pre}.
\label{sec:ar_hyperparams}
\begin{table}[H]
    \caption{Fine-tuning and Baseline Hyperparameters}
    \centering
    \resizebox{\linewidth}{!}{
    \begin{tabular}{c|cccccc}
         Model & Batch Size & Learning Rate & Weight Decay & Dropout & \\
         \hline
         FNO & 16 & 1E-3 & 1E-6 & 0.0 \\
         PICL FNO & 16 & 1E-2 & 1E-6 & 0.0\\
    \end{tabular}}
    \label{tab:1d_ar_train}
\end{table}
\begin{table}[H]
    \caption{Pretraining Hyperparameters}
    \centering
    \resizebox{\linewidth}{!}{
    \begin{tabular}{c|cccccc}
         Model & Batch Size & Learning Rate & Weight Decay & Dropout & $\tau$ \\
         \hline
         Pretrain FNO & 1E-2 & 1E-8 & 10 & 0.0 & 1\\
    \end{tabular}}
    \label{tab:1d_ar_pre}
\end{table}

\subsection{2D Hyperparameters}
For all 2D experiments, we used a 2D FNO with hidden width of 48 and 4 modes and 4 layers
that was trained for 500 epochs.
Pretraining was done for 500 epochs in the fixed-future experiment and 100 for the autoregressive rollout experiment, with 5000 samples from each equation.
For $\tau = \texttt{mean}$, we take set $\tau$ to the mean distance value for each batch.
\subsubsection{Fixed-Future Hyperparameters}
\label{sec:ff_hyperparams}
Hyperparameters for finetuning and baseline training are given in table \ref{tab:2d_ff_train}, and for pretraining in table \ref{tab:2d_ff_pre}.
\begin{table}[H]
    \caption{Fine-tuning and Baseline Hyperparameters}
    \centering
    \resizebox{\linewidth}{!}{
    \begin{tabular}{c|cccccc}
         Model & Batch Size & Learning Rate & Weight Decay & Dropout & \\
         \hline
         FNO & 32 & 1E-2 & 1E-7 & 0.0 \\
         PICL FNO & 32 & 1E-2 & 1E-7 & 0.0 \\
    \end{tabular}}
    \label{tab:2d_ff_train}
\end{table}
\begin{table}[H]
    \caption{Pretraining Hyperparameters}
    \centering
    \resizebox{\linewidth}{!}{
    \begin{tabular}{c|ccccccc}
         Model & Batch Size & Learning Rate & Weight Decay & Dropout & $\tau$ \\
         \hline
         Pretrain FNO & 256 & 1E-2 & 1E-7 & 0.00 & Mean \\
    \end{tabular}}
    \label{tab:2d_ff_pre}
\end{table}

\subsubsection{Autoregressive Rollout Hyperparameters}
\label{sec:ar_hyperparams}
Hyperparameters for finetuning and baseline training are given in table \ref{tab:2d_ar_train}, and for pretraining in table \ref{tab:2d_ar_pre}.
\begin{table}[H]
    \caption{Fine-tuning and Baseline Hyperparameters}
    \centering
    \resizebox{\linewidth}{!}{
    \begin{tabular}{c|cccccc}
         Model & Batch Size & Learning Rate & Weight Decay & Dropout & \\
         \hline
         FNO & 16 & 1E-3 & 1E-6 & 0.0 \\
         PICL FNO & 16 & 1E-2 & 1E-6 & 0.0\\
    \end{tabular}}
    \label{tab:2d_ar_train}
\end{table}
\begin{table}[H]
    \caption{Pretraining Hyperparameters}
    \centering
    \resizebox{\linewidth}{!}{
    \begin{tabular}{c|cccccc}
         Model & Batch Size & Learning Rate & Weight Decay & Dropout & $\tau$ \\
         \hline
         Pretrain FNO & 1E-2 & 1E-8 & 10 & 0.0 & 1\\
    \end{tabular}}
    \label{tab:2d_ar_pre}
\end{table}

\section{Passthrough}
\label{app:passthrough_tsne}

\begin{figure}[H]
    \centering
    \includegraphics[width=\linewidth]{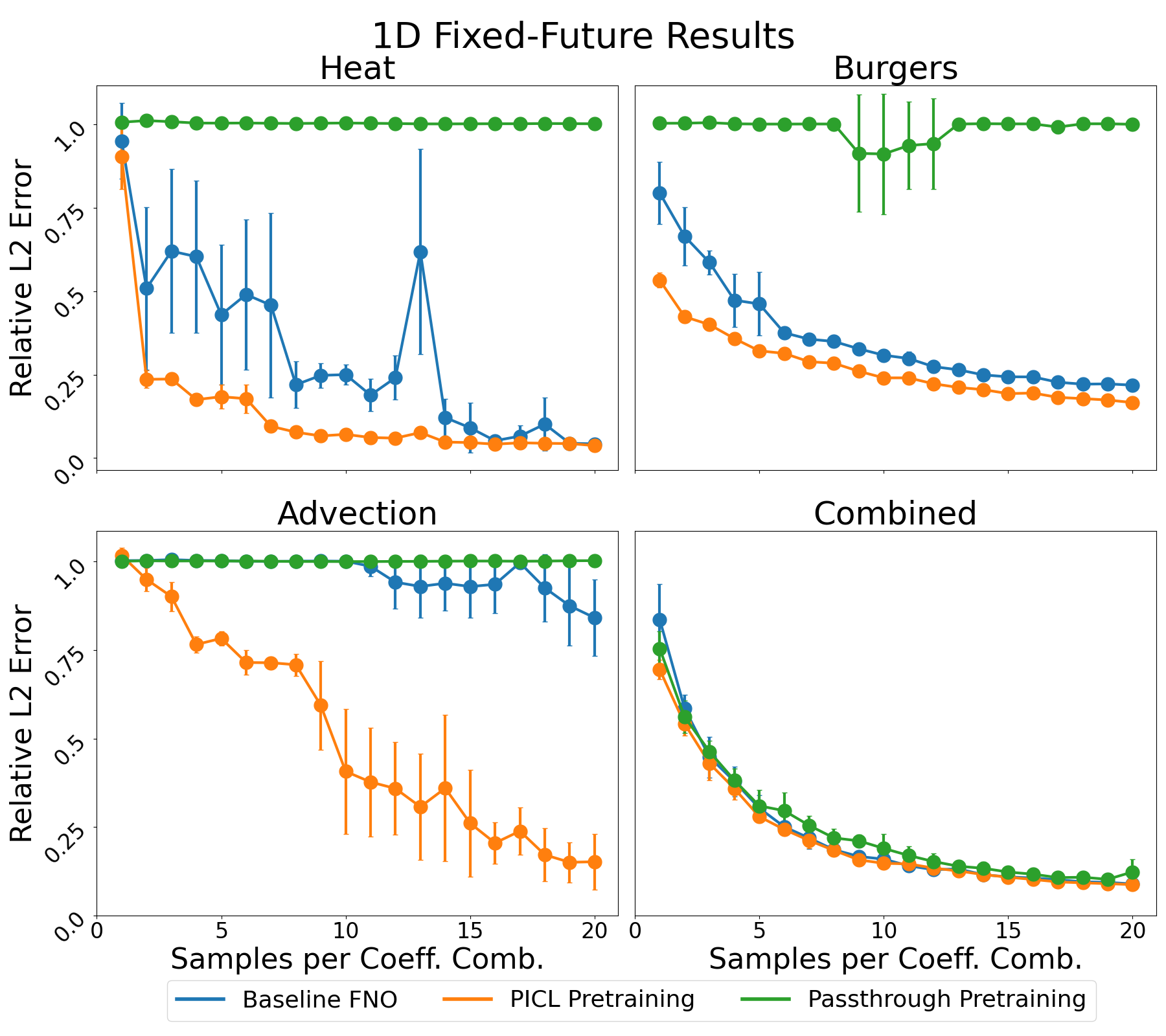}
    \caption{1D comparison of fixed-future performance between FNO, FNO pretrained using PICL, and FNO pretrained using passthrough.}
    \label{fig:passthrough_1d_ff_comp}
\end{figure}

\begin{figure}[H]
    \centering
    \includegraphics[width=\linewidth]{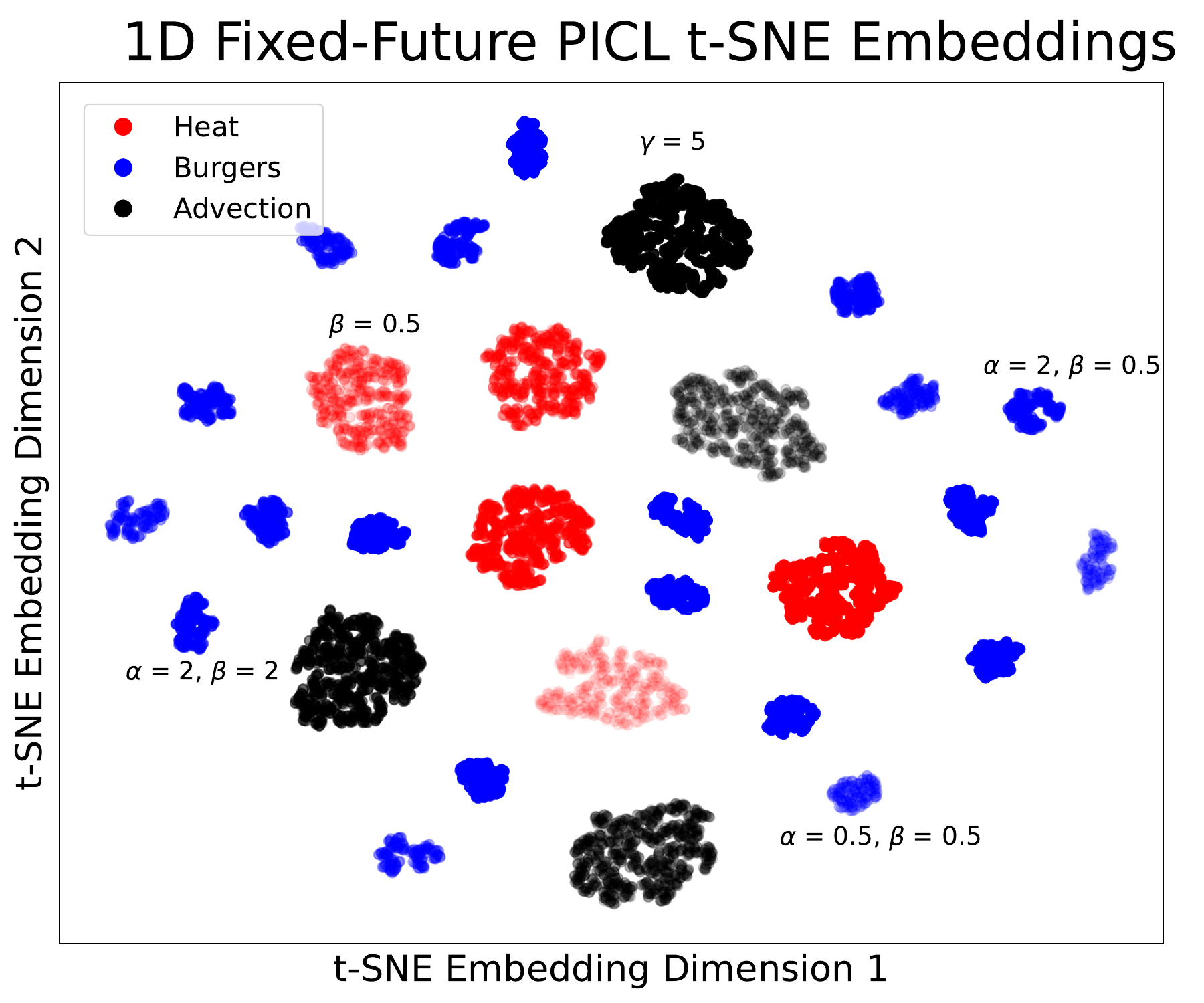}
    \caption{t-SNE of latent embeddings after passthrough pretraining. We see clear clusters for each combination of operator coefficients.}
    \label{fig:passthrough_tsne}
\end{figure}

For passthrough pretraining, we have $d_{system}(u_i) = u_i$ and $d_{update}(u_i) = G(u_i)$.
When we exclude physics information from our pretraining loss, we see the model is unable to learn during fine-tuning in figure \ref{fig:passthrough_1d_ff_comp}.
In the t-SNE plot, we see very neat clusters for each combination of equation coefficients in figure \ref{fig:passthrough_tsne}.
While this shows excellent structure, it does not match intuition that diffusion dominated Burgers' systems behave more similarly to Heat systems than advection dominated Burgers' systems.


\section{Autoregressive Results}
\label{app:ns_ar}
\begin{figure}[H]
    \centering
    \includegraphics[width=\linewidth]{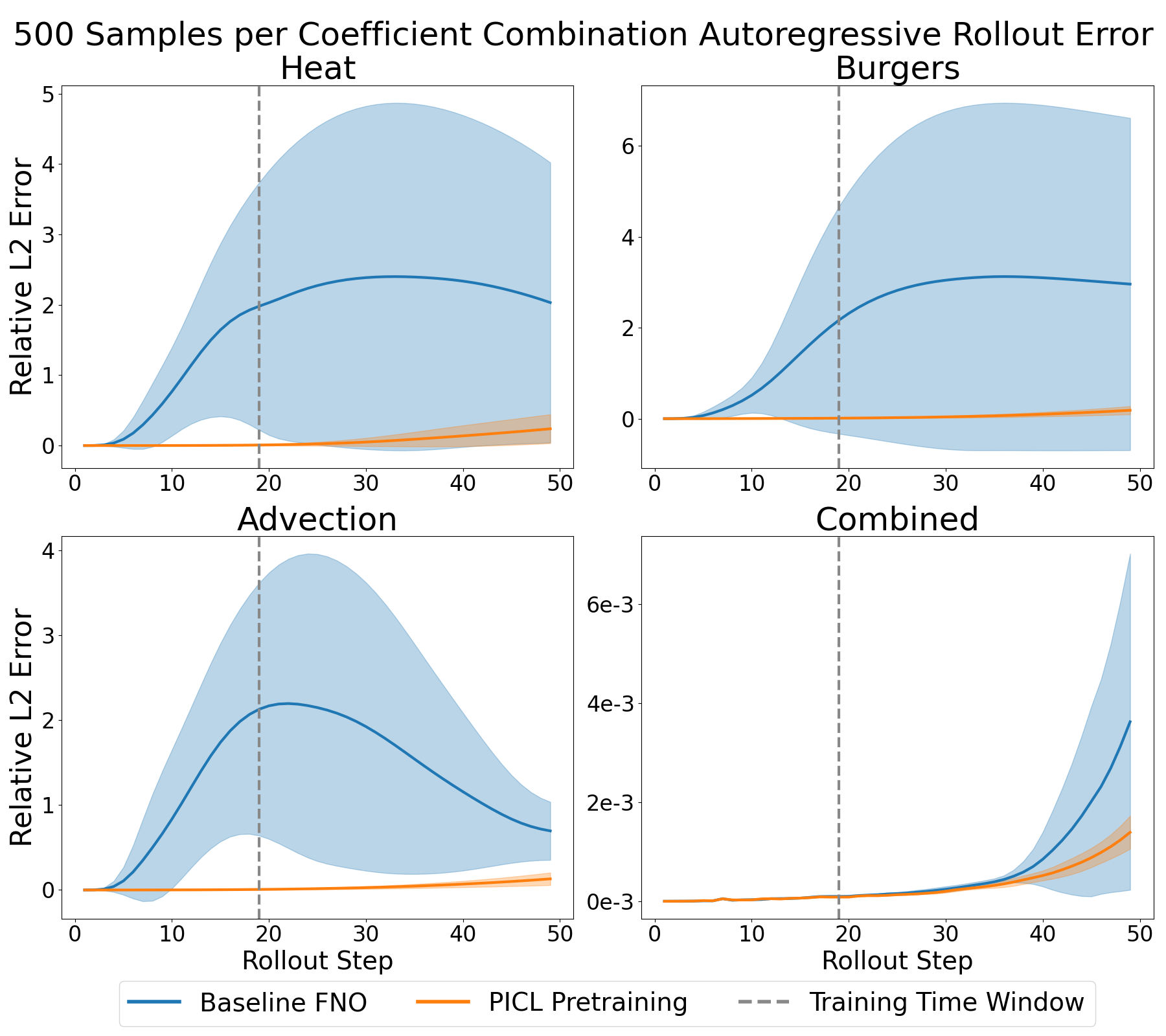}
    \caption{Comparison of autoregressive rollout performance between FNO and FNO pretrained using PICL.}
    \label{fig:rollout}
\end{figure}

\begin{figure}[H]
    \centering
    \includegraphics[width=\linewidth]{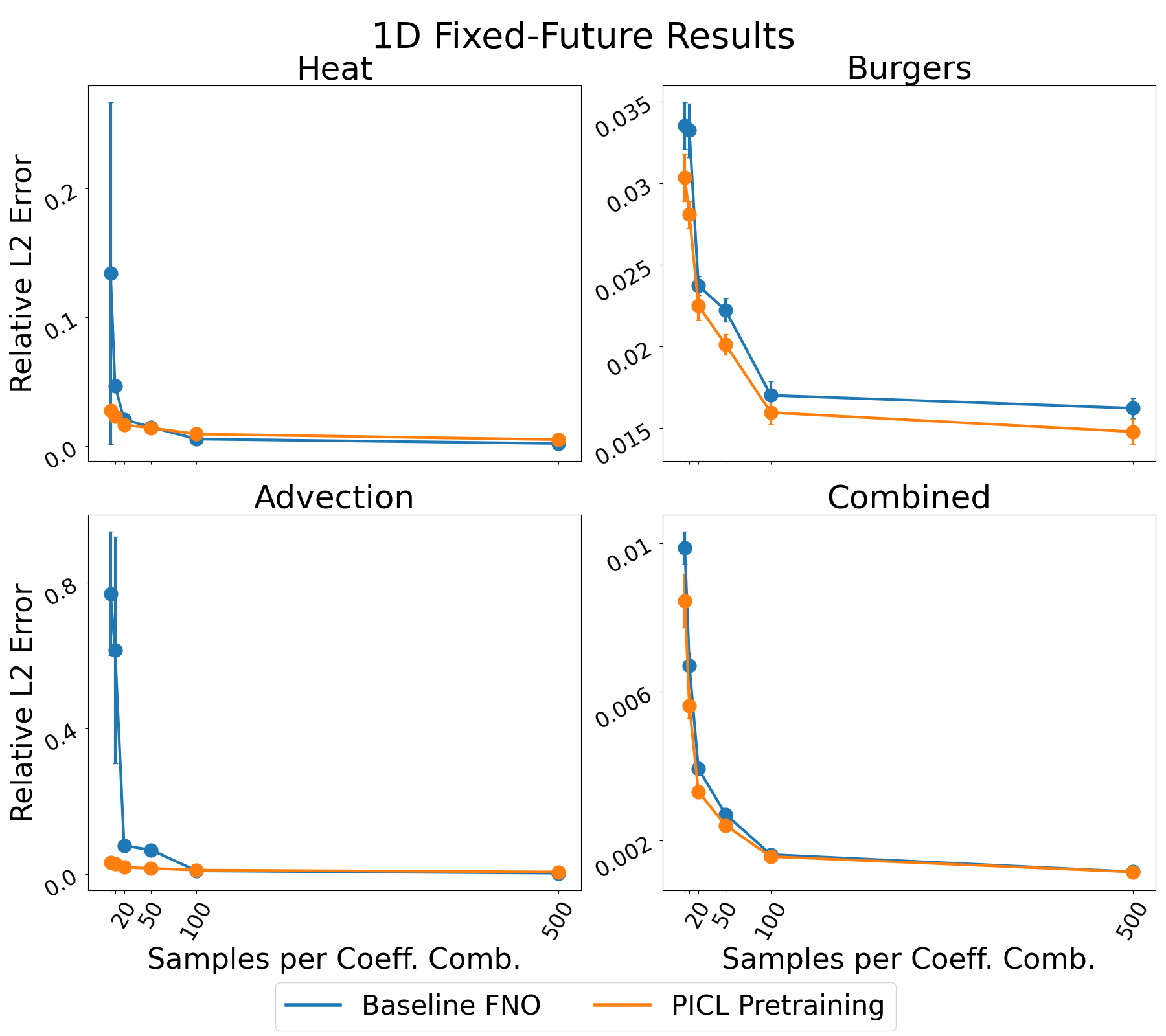}
    \caption{Comparison of autoregressive rollout performance between FNO and FNO pretrained using PICL.}
    \label{fig:1d_next_step_pred}
\end{figure}

Looking further at our next-step training and autoregressive rollout results, we see that baseline FNO is more unstable than when pretrained with PICL.
For our individual data sets, seen in figure \ref{fig:rollout}, error accumulates accumulates significantly before our training time window, and after our training time window on the combined data set.
Rollout is unstable for baseline training despite comparable performance in next-step predictive accuracy, seen in figure \ref{fig:1d_next_step_pred}.

\end{document}